# Dual Architecture Fusion Vehicle Lidar 3D Target Detection

Muyun Li
*WHU*
*EIS*
Wuhan, China
muyunli07@outlook.com

Weizheng Jin*
*WHU*
*EIS*
Wuhan, China
jwz@whu.edu.cn

Yuan Sun
*WHU*
*EIS*
Wuhan, China
3396897122@qq.com

***Abstract*—In this paper, a 3D backbone network (SpconvTransformer) based on the fusion of convolutional architecture branch and Transformer architecture branch is proposed. The convolution branch uses 3D sparse convolution and submanifold convolution to learn local voxel features and semantic information, and the Transformer branch uses the self-attention mechanism to model the remote relationship between voxels, uses the broad receptive field to capture rich context information, and complementing the advantages of the two architectures by fusing the local features extracted from the two branches and the global features solves the problem of limited receptive fields of 3D convolutional neural networks. The experimental results show that the average accuracy of SpconvTransformer in 3D object detection and bird's-eye view detection tasks reaches 81.45% and 88.09%, respectively, and the average directional similarity reaches 91.10%, which is significantly better than other existing algorithms.**

## I. Introduction

LiDAR-based 3D object detection technology is widely utilized in autonomous driving systems [1]. In recent years, the field has advanced rapidly, with sparse voxel Transformer backbones such as DSVT [2] and unified (often multi-modal) Transformer backbones for BEV representation such as UniTR [3] achieving notable performance gains. The Transformer architecture [4] can effectively capture global features and rich contextual information through its self-attention mechanism. Drawing on the idea of VoTr [5] that employs Transformer to establish long-range dependencies between voxels, this paper proposes a voxel feature extraction network named SpconvTransformer, which integrates a sparse convolution architecture with the Transformer architecture [4, 5]. The new design fully exploits the capability of convolutional neural networks in modeling local voxel features, as well as the ability of Transformer's self-attention mechanism to model global voxel features. By fusing global features containing contextual information with local features that preserve detailed information, SpconvTransformer achieves complementary advantages in voxel feature extraction. Furthermore, recent advances in 2024 and 2025, such as Local Sparse Transformers [6] and SFMNet [7], have strongly validated that integrating sparse convolutions with focal or local attention mechanisms is a highly effective trend for balancing local structural details and large-scale 3D perception. Compared with recent dynamic set-based sparse voxel attention schemes [2], our dual-branch fusion strategy provides a more direct way to balance local details and global context. Experiments demonstrate that SpconvTransformer exhibits excellent performance in point cloud 3D object detection tasks. This strategy also aligns with the recent success of center-based detectors and robust multi-modal fusion frameworks in complex 3D scenes [8, 9].

## II. Basic Principles of the 3D Object Detection Algorithm Based on Multi-branch Feature Fusion

### *A. Algorithm Pipeline*

The overall pipeline of the SpconvTransformer point cloud 3D object detection algorithm is shown in Figure 1, which can be divided into the following three stages: voxel feature encoding, the 3D backbone network based on SpconvTransformer, and the Region Proposal Network (RPN).

The algorithm first partitions the raw point cloud scene into regular voxels and assigns the point data to the corresponding voxels. Each voxel retains at most 5 points after random point sampling. A mean voxel feature encoding module calculates the average of the coordinate information of all points within each voxel to serve as the initial feature of the voxel. The processed per-voxel features are then fed separately into a 3D sparse-convolution-based backbone network and a Voxel-Transformer-based backbone network, which extract local voxel features and long-range voxel features via convolutional neural networks and the self-attention mechanism, respectively. Subsequently, the features extracted from the two branches are fused via concatenation, aggregating the local detail information and the contextual information of the voxels. Finally, a Region Proposal Network (RPN) is employed to perform object classification, bounding-box regression, and orientation prediction.

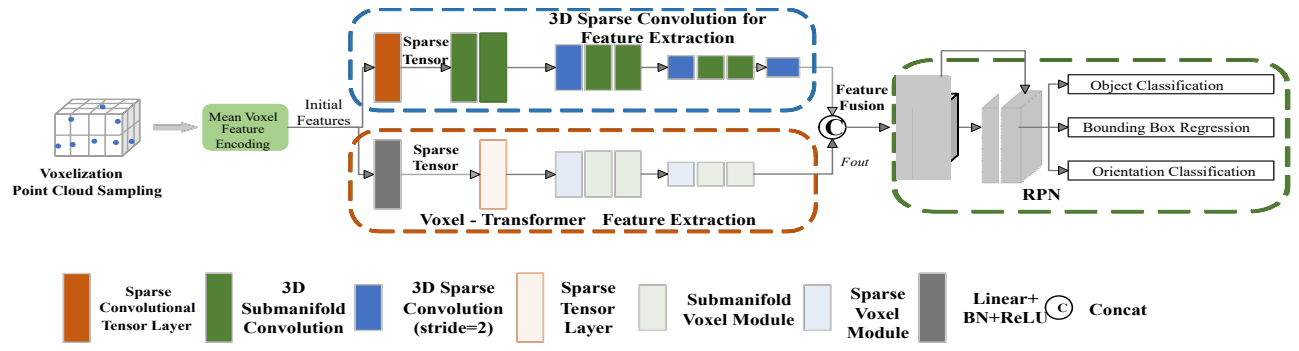

Figure 1. Overall Network Architecture.

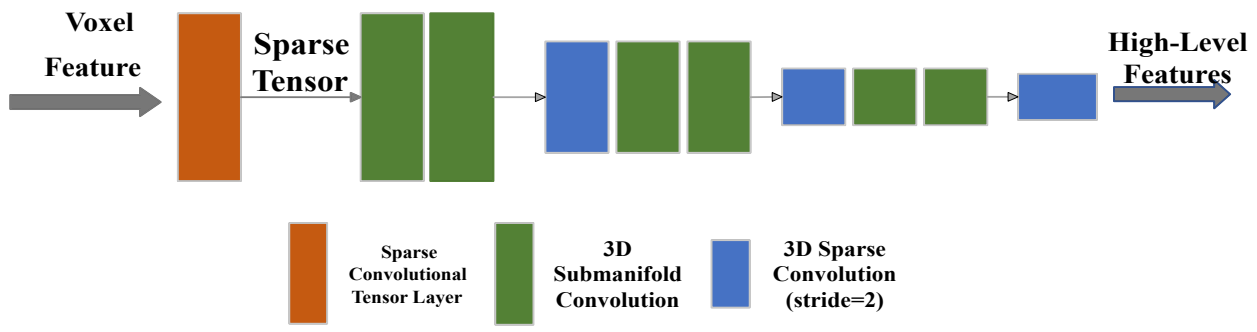

Figure 2. Sparse Convolution-based Feature Extraction Module

## B. *Sparse Convolution Feature Extraction Module*

Convolutional Neural Networks (CNNs) are capable of efficiently extracting local spatial information. The voxel data input to the algorithm is highly regular, making it suitable for learning high-level feature representations using CNNs. However, as VoxelNet has demonstrated, 3D CNNs can impose a significant computational burden [10].

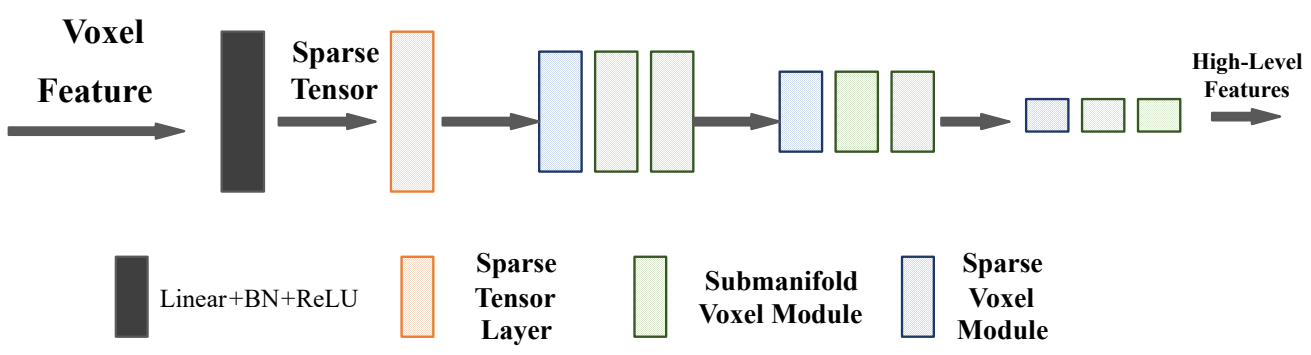

Figure 3. Transformer-based Feature Extraction Module.

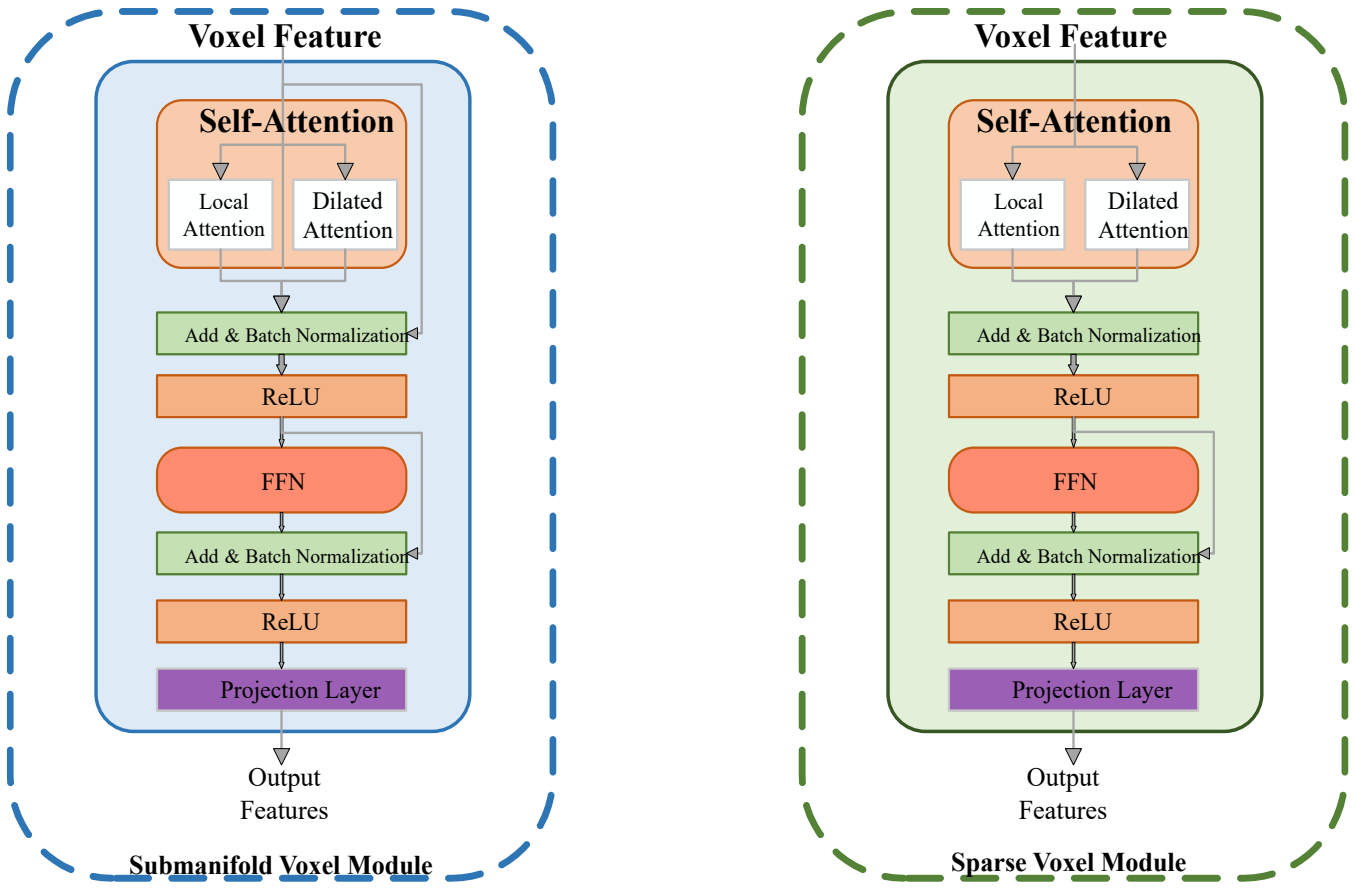

Figure 4. Principle of Submanifold and Sparse Voxel Modules.

As shown in Figure 2, this module first utilizes a sparse convolutional tensor layer to convert the initial voxel features into a sparse four-dimensional tensor representation. It then employs submanifold convolution and 3D sparse convolution to deepen the network and perform down-sampling. The three-dimensional space undergoes eight-fold down-sampling while advanced features are extracted. The algorithm also adopts dimension compression, further condensing the down-sampled 3D space into a dense 2D bird's-eye view (BEV) representation. This leverages the advantage of the bird's-eye view, which is less affected by occlusion and is more conducive to object recognition as well as bounding-box localization and regression [3].

## C. *Transformer Feature Extraction Module*

Long-range dependencies also exist among voxels in three-dimensional space. Drawing an analogy to the self-attention mechanism in the Transformer architecture, which models long-range relationships between pixels to capture contextual information, the Transformer architecture can be applied for voxel feature extraction.

For 3D spaces containing a large number of non-empty voxels, the Transformer feature extraction module in this algorithm employs a submanifold voxel module and a sparse voxel module to extract features from non-empty and empty voxels, respectively. As shown in Figure 3, the algorithm utilizes three sparse voxel modules to achieve eight-fold downsampling of the scene. Furthermore, two submanifold voxel modules are used after each sparse voxel module to maintain the stability of non-empty voxel positions, thereby preserving the original 3D spatial structure. A comparison with Figure 2 reveals that the basic architecture of the two feature extraction modules is similar, but they are implemented using different operators.

As shown in Figure 4, both the sparse voxel module and the submanifold voxel module are built upon the local attention mechanism and the dilated attention mechanism. The local attention mechanism focuses on the neighboring region of a query voxel, thereby preserving more detailed information. The dilated attention mechanism progressively increases the search stride, enabling the acquisition of a larger receptive field with a smaller number of voxels. The difference between the two modules lies in the fact that the submanifold voxel module additionally incorporates a skip-connected "residual structure," which fuses the initial voxel features with the voxel features processed by the self-attention mechanism. This design ensures that the positions of empty voxels remain unchanged as the network deepens.

## D. *SpconvTransformer Backbone Network*

The 3D sparse convolution-based feature extraction module and the Transformer-based feature extraction module process the encoded per-voxel features. The former efficiently extracts local features from regular voxels while learning rich semantic information from the scene, whereas the latter branch captures extensive contextual information from the scene and obtains a larger receptive field while preserving spatial detail. Therefore, this paper combines the two architectures and proposes a voxel feature extraction backbone network named SpconvTransformer, which achieves the fusion of local and global information. It fully leverages the complementary modeling capabilities of the two distinct architectures to extract higher-quality voxel features for point cloud 3D object detection tasks.

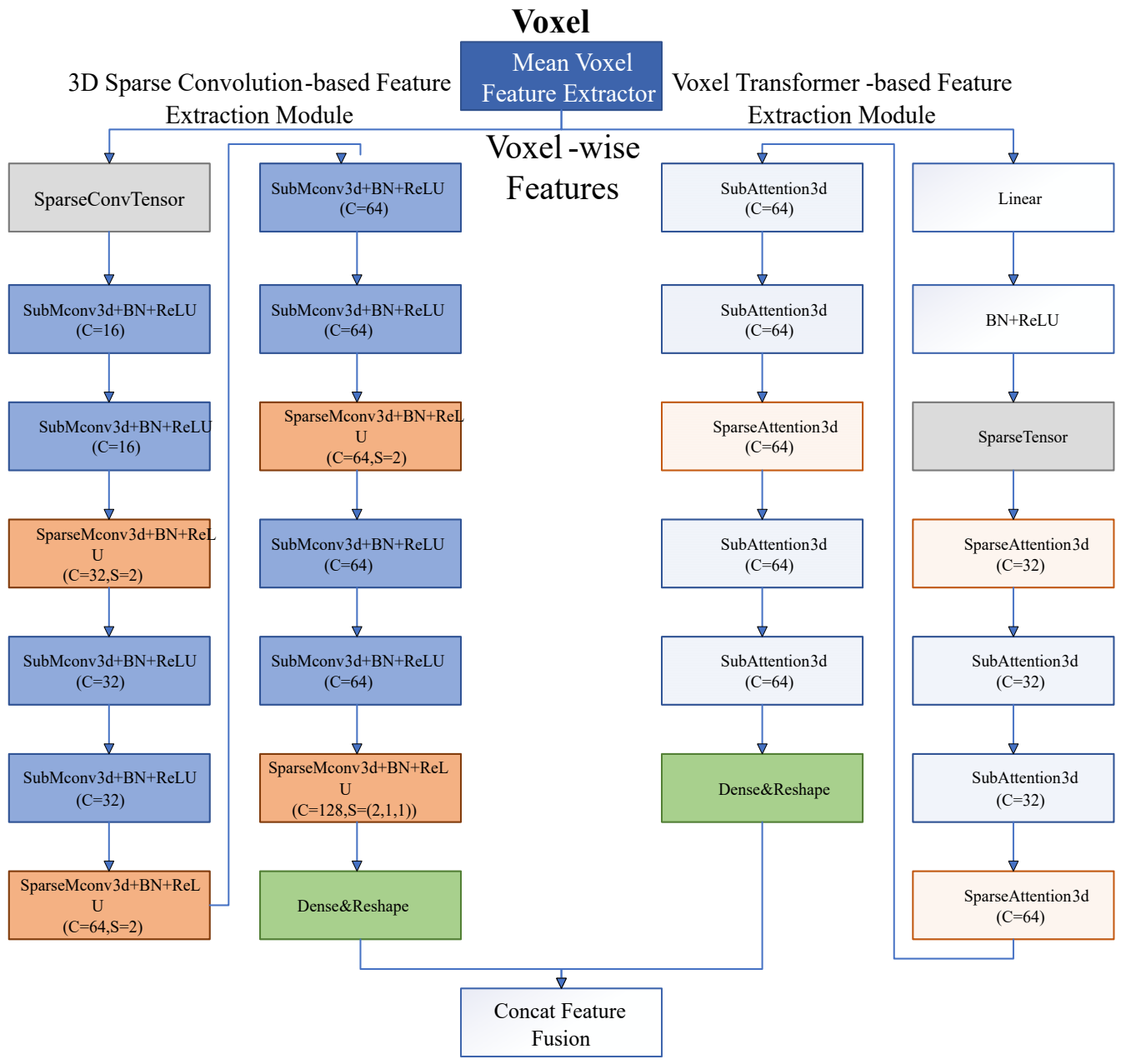

Figure 5. Multi-Branch Fusion-Based 2D Backbone Network.

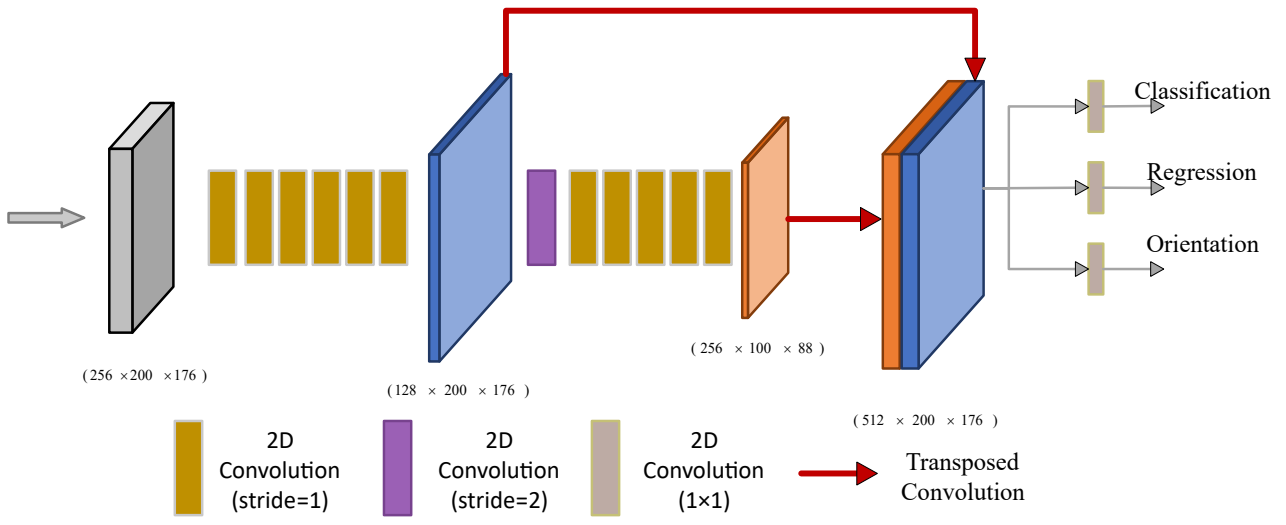

Figure 6. Region Proposal Network (RPN) Architecture

The structure of the SpconvTransformer backbone network is illustrated in Figure 5. For clarity, Figure 5 uses SubMAttention3d and SparseAttention3d to denote the submanifold voxel module and sparse voxel module in the Transformer architecture, respectively, and uses SubMConv3d and SparseConv3d to represent 3D submanifold convolution and 3D sparse convolution. The entire feature extraction network takes voxelized point clouds as input. The initial voxel features are fed into the convolutional architecture and the Transformer architecture separately. The orange rectangular blocks perform down-sampling operations on the sparse data, while the light blue rectangular blocks further learn feature representations of the voxels while maintaining data sparsity. Both branches perform eight-fold down-sampling of the 3D scene through three 3D sparse convolutional layers and three sparse voxel layers, respectively, followed by a densification operation and dimension reshaping. Specifically, the data on the left branch is finally compressed into a 4D tensor of size 128 × 2 × 200 × 176, where 128 is the number of feature channels, and 2, 200, 176 are the spatial dimensions. The data on the right branch is compressed into a 4D sparse tensor of size 64 × 5 × 200 × 176, where 64 is the number of feature channels, and 5, 200, 176 are the spatial dimensions along the three axes.

Due to the reduced occlusion in bird's-eye view representations, the algorithm multiplies and merges the channel dimension with the z-axis dimension, thereby converting the sparse 3D spaces from both branches into dense 2D bird's-eye views (BEVs). The features extracted by the convolutional branch contain higher-level semantic information and precise local details, while those from the Transformer branch encompass richer contextual information. To leverage the complementary strengths of both, the algorithm employs a channel-wise concatenation method to fuse the two feature streams, ultimately producing a 2D bird's-eye view representation with dimensions of 576 × 200 × 176.

### E. 2D Backbone Network Based on Multi-Branch Fusion

The Region Proposal Network (RPN) is an indispensable part of a 3D object detector, consisting mainly of a 2D backbone network and a detection head. After voxel data is processed by SpconvTransformer to extract high-level feature representations and converted into a dense 2D bird's-eye view (BEV), a simplified RPN can be used to achieve 3D object classification, bounding box regression, and orientation prediction. Given the substantial computational cost of SpconvTransformer, this algorithm employs a lightweight 2D backbone network, whose structure is shown in Figure 6. For clarity, this paper uses Conv2d(Cin, Cout, K, S) to denote a standard 2D convolution with an input channel count of Cin, an output channel count of Cout, a kernel size of K, and a stride of S; ConvTranspose2d(Cin ,Cout,S) SpconvTransformer denotes a 2D transposed convolution with an input channel count of Cin, an output channel count of Cout, and an upsampling stride of S.

The Region Proposal Network (RPN) takes the bird's-eye view (BEV) features extracted by SpconvTransformer as input. It first employs a Conv2d(256, 256, 3, 1) layer and five Conv2d(128, 128, 3, 1) layers to extract high-level features at the same scale. Subsequently, a Conv2d(128, 256, 3, 2) layer with a stride of 2 is used for down-sampling, followed by five consecutive Conv2d(256, 256, 3, 1) layers to enrich the semantic information of the features. After obtaining features at two different scales, transposed convolution layers—ConvTranspose2d(128, 256, 1) with stride 1 and ConvTranspose2d(128, 256, 2) with stride 2—are applied to adjust the feature dimensions. The processed features are then concatenated to produce the final fused feature output. Finally, three 1×1 2D convolutional layers—Conv2d(512, 2, 1, 1), Conv2d(512, 14, 1, 1), and Conv2d(512, 4, 1, 1)—are utilized to predict the object category, bounding box, and orientation state, respectively.

## III. Experimental Design and Results Analysis

### A. Experimental Platform and Training

Due to the substantial computational cost of the Transformer architecture in 3D scenarios and its high demand for GPU memory during training, the Adam optimizer is employed in this algorithm to train the model for 100 epochs. The batch size is set to 1, meaning a total of 371,200 iterations are required for the 3,712 training samples, with one sample processed per iteration. Training the model on a single GTX 1080Ti GPU takes approximately 65 hours.

### B. Performance Metric Comparison

To validate the effectiveness of the algorithm in 3D object detection and bird's-eye view (BEV) object detection tasks, experiments are conducted in this section involving model training and evaluation on the public KITTI dataset. The dataset is split into 3,712 training samples for model training and 3,769 validation samples for calculating performance metrics. As most existing algorithms use the AP_R11 metric on the validation set for comparison, for a fair and objective evaluation of the proposed algorithm, the performance metrics in this subsection also adopt the AP_R11 detection accuracy at an IoU threshold of 0.7. Table 1 presents the 3D and BEV object detection performance of various state-of-the-art algorithms on the KITTI validation set. The best performance for each task is highlighted in bold. It should be noted that the two-stage detectors listed in the table all use multimodal data input ("color image + LiDAR point cloud"), while the single-stage detectors use only LiDAR point clouds as input. "Mod." in the table is the abbreviation for the moderate difficulty level. The best performance metrics for each task are emphasized in bold.

As can be seen from Table 1, the proposed algorithm achieves highly competitive performance in both 3D object detection and bird's-eye view (BEV) object detection tasks, with average precision reaching 81.45% and 88.09%, respectively. Moreover, its detection accuracy significantly outperforms other algorithms across the easy, moderate, and hard difficulty levels.

The single-stage detection algorithms listed in the table—VoxelNet, SECOND, and VoTr-SSD—are based on 3D convolution architecture, 3D sparse convolution architecture, and Transformer architecture, respectively. The comparison shows that employing the SpconvTransformer proposed in this work as the 3D backbone network architecture enables higher-accuracy point cloud object detection.

TABLE 1. PERFORMANCE OF DIFFERENT ALGORITHMS ON THE KITTI VALIDATION SET FOR VEHICLE DETECTION (IoU=0.7).

| Type | Method | 3D Detection (IoU = 0.7) | | | | BEV Detection (IoU = 0.7) | | | | FPS |
|---|---|---|---|---|---|---|---|---|---|---|
| | | *Easy* | *Mod.* | *Hard* | *mAP* | *Easy* | *Mod.* | *Hard* | *mAP* | |
| *2-stage* | ***MV3D[11]*** | 71.29 | 62.68 | 56.56 | 63.51 | 86.55 | 78.10 | 76.67 | 80.44 | 3 |
| | ***AVOD-FPN[12]*** | 84.41 | 74.44 | 68.65 | 75.83 | N/A | N/A | N/A | N/A | 10 |
| | ***F-PointNet[13]*** | 83.76 | 70.92 | 63.65 | 72.78 | 88.16 | 84.02 | 76.44 | 82.87 | 6 |
| *1-stage* | ***VoxelNet[10]*** | 81.97 | 65.46 | 62.85 | 70.09 | 89.60 | 84.81 | 78.57 | 84.33 | 4 |
| | ***SECOND[14]*** | 87.43 | 76.48 | 69.10 | 77.67 | 89.96 | 87.07 | 79.66 | 85.56 | 25 |
| | ***SECOND-S*** | 85.50 | 75.04 | 68.78 | 76.44 | 89.79 | 86.20 | 79.55 | 85.18 | 40 |
| | ***PointPillars[15]*** | 86.13 | 77.03 | 72.43 | 78.53 | 89.93 | 86.92 | 84.97 | 87.27 | **62** |
| | ***VoTr-SSD[5]*** | 87.86 | 78.27 | 76.93 | 81.02 | N/A | N/A | N/A | N/A | 15 |
| | ***Algorithm We use*** | **88.50** | **78.33** | **77.53** | **81.45** | **90.07** | **87.58** | **86.63** | **88.09** | 7 |

TABLE 2. COMPARISON OF AVERAGE ORIENTATION SIMILARITY (AOS) PERFORMANCE ON THE KITTI VALIDATION SET (%).

| Method | Modality | AOS (IoU = 0.7) | | | mAOS |
|---|---|---|---|---|---|
| | | *Easy* | *Moderate* | *Hard* | |
| ***SubCNN*** | RGB | 90.61 | 88.43 | 78.63 | 85.89 |
| ***AVOD-FPN*** | RGB+LiDAR | 89.95 | 87.13 | 79.74 | 85.61 |
| ***SECOND*** | LiDAR | 87.84 | 81.31 | 71.95 | 80.37 |
| ***PointPillars*** | LiDAR | 90.19 | 88.76 | 86.38 | 88.44 |
| ***Algorithm we use*** | LiDAR | **95.30** | **89.36** | **88.65** | **91.10** |
| ***Improvement*** | N/A | +4.69 | +0.60 | +2.27 | +2.66 |

## C. Ablation Study

An ablation study is conducted using two performance evaluation metrics—the classic AP_R11 and the state-of-the-art AP_R40—to demonstrate the contribution of each module in the algorithm. Table 2 and Table 3 show the impact of individual modules on algorithm performance when using AP_R11 and AP_R40 as the evaluation metrics, respectively. AP_R11, as a classic evaluation method, allows for comparison with existing publicly reported algorithms. AP_R40, adopted by KITTI as a more rigorous metric, can more accurately reflect changes in network performance. As shown in Table 2 and Table 3, "Spconv" denotes using only the convolutional neural network architecture, where the backbone is built with 3D sparse convolution and 3D submanifold convolution. Correspondingly, "Transformer" in the table indicates using only the Transformer architecture, which extracts voxel features via the self-attention mechanism. "SpconvTransformer+Sum" refers to using the dual-architecture fused 3D backbone for feature extraction, where features from the Spconv and Transformer branches are fused via element-wise summation. "SpconvTransformer+ Concat" denotes feature fusion using channel-wise concatenation. The last two rows of the table present the performance improvement of the proposed algorithm relative to the two single-architecture solutions.

As shown in Table 2, when using a single architecture for voxel feature extraction, the method based on the Spconv architecture achieves slightly higher average precision in both 3D object detection and bird's-eye view (BEV) object detection compared to the Transformer architecture. When the two architectures are fused via element-wise summation, the detection accuracy for easy-level tasks in both 3D and BEV object detection is significantly improved, with moderate and slight improvements also observed for moderate and hard levels. Directly concatenating the feature outputs of the two branches (Concat) yields better performance than adjusting the feature outputs of both branches to the same number of channels followed by element-wise summation (Sum).

The reason is that the features extracted by the two architectures have low similarity. Simply adding the local features from the convolutional architecture and the global features from the Transformer architecture element-wise may cause the information in the fused features to become mixed, where useful information can become blurred, leading to information loss. In contrast, concatenation preserves all channel information, allowing the neural network to learn during training how to effectively fuse the useful parts of both features, thereby avoiding the information loss caused by direct feature addition.

From Table 2, it can be seen that when using Concat to fuse features extracted by the two architectures, the average precision for 3D object detection and BEV object detection reaches 81.45% and 88.09%, respectively. Compared to the single-architecture methods, the average precision for 3D object detection is improved by 0.84% and 1.58%, and for BEV detection by 0.75% and 1.06%. This indicates that the feature extraction capability of the Transformer architecture is slightly lower than that of the Spconv architecture. However, the self-attention mechanism with a larger receptive field can capture rich contextual information in the scene, thereby compensating for the contextual information lacking in the Spconv architecture while also expanding the overall receptive field of the network, achieving complementary advantages.

Under the more precise AP_R40 evaluation metric, the detector based on the dual-architecture Concatenation fusion achieves average precisions of 83.81% and 89.84% in 3D and bird's-eye view (BEV) object detection tasks, respectively. The performance improvement is particularly notable for 3D object detection, with average precision increasing by 1.01% and 3.00% compared to the single-architecture Spconv and Transformer methods, respectively. In contrast, the dual-architecture Sum fusion method yields a more significant improvement for BEV object detection. As this study focuses primarily on 3D object detection in point cloud scenes, the Concat fusion scheme is ultimately selected for the proposed SpconvTransformer algorithm to integrate the dual architectures.

TABLE 3. AVERAGE PRECISION AP_R11 (%) FOR 3D AND BIRD'S-EYE VIEW (BEV) OBJECT DETECTION.

| **Method** | **3D Detection (IoU = 0.7)** | | | | **BEV Detection (IoU = 0.7)** | | | |
|---|---|---|---|---|---|---|---|---|
| | ***Easy*** | ***Mod.*** | ***Hard*** | ***mAP*** | ***Easy*** | ***Mod.*** | ***Hard*** | ***mAP*** |
| ***Spconv*** | 87.58 | 77.90 | 76.36 | 80.61 | 89.65 | 87.26 | 85.12 | 87.34 |
| ***Transformer*** | 87.16 | 77.17 | 75.29 | 79.87 | 89.63 | 86.94 | 84.52 | 87.03 |
| ***SpconvTranformer+Sum*** | 88.28 | 77.95 | 76.86 | 81.03 | 89.97 | 87.44 | 86.25 | 87.89 |
| ***SpconvTranformer+ Concat*** | **88.50** | **78.33** | **77.53** | **81.45** | **90.07** | **87.58** | **86.63** | **88.09** |
| ***IMP(Spconv)*** | +0.92 | +0.43 | +1.17 | +0.84 | +0.42 | +0.32 | +1.51 | +0.75 |
| ***IMP(Transformer)*** | +1.34 | +1.16 | +2.24 | +1.58 | +0.44 | +0.64 | +2.11 | +1.06 |

TABLE 4. AVERAGE PRECISION AP_R40 (%) FOR 3D AND BIRD'S-EYE VIEW (BEV) OBJECT DETECTION.

| **Method** | **3D Detection (IoU = 0.7)** | | | | **BEV Detection (IoU = 0.7)** | | | |
|---|---|---|---|---|---|---|---|---|
| | ***Easy*** | ***Mod.*** | ***Hard*** | ***mAP*** | ***Easy*** | ***Mod.*** | ***Hard*** | ***mAP*** |
| ***Spconv*** | 89.75 | 80.81 | 77.83 | 82.80 | 92.39 | 88.28 | 87.08 | 89.25 |
| ***Transformer*** | 88.20 | 78.55 | 75.69 | 80.81 | 92.30 | 88.08 | 85.47 | 88.62 |
| ***SpconvTranformer+Sum*** | **90.84** | 80.98 | 78.50 | 83.44 | **94.64** | **88.65** | 87.93 | **90.41** |
| ***SpconvTranformer+ Concat*** | 90.65 | **81.59** | **79.20** | **83.81** | 92.81 | 88.61 | **88.11** | 89.84 |
| ***IMP(Spconv)*** | +0.90 | +0.78 | +1.37 | +1.01 | +0.42 | +0.33 | +1.03 | +0.59 |
| ***IMP(Transformer)*** | +2.45 | +3.04 | +3.51 | +3.00 | +0.51 | +0.53 | +2.64 | +1.22 |

## IV. CONCLUSION

To more effectively extract voxel features, this paper proposes a novel architecture named SpconvTransformer, which fuses sparse convolution and Transformer. By integrating the local features extracted by convolutional layers with the global features captured by the self-attention mechanism, the issue of limited receptive field in the model is addressed. When this architecture is employed as a 3D backbone network for point cloud object detection tasks in autonomous driving scenarios, it enables high-precision 3D and bird's-eye view object detection. The competitive performance of the new architecture can better serve both educational and scientific research endeavors.

## ACKNOWLEDGEMENT

This work was supported in part by the Natural Science Foundation of Hubei Province under Grant 2020CFA053.(Corresponding author: Weizheng Jin.)